\documentclass[letterpaper]{article} 
\usepackage[preprint]{aaai2027}
\usepackage[hyphens]{url}  
\usepackage{graphicx} 
\usepackage{natbib}  
\usepackage{caption} 
\usepackage{amsmath}
\usepackage{algorithm}
\usepackage{algorithmic}

\usepackage{newfloat}
\usepackage{listings}
\DeclareCaptionStyle{ruled}{labelfont=normalfont,labelsep=colon,strut=off} 
\floatstyle{ruled}
\newfloat{listing}{tb}{lst}{}
\floatname{listing}{Listing}

\usepackage{booktabs}

\usepackage[table]{xcolor}
\definecolor{lightblue}{RGB}{222,235,255}

\title{Skeleton-Guided Progressive Test-Time Adaptation for \\ Thin Curvilinear Structures}
\author{
    Boa Jang\textsuperscript{\rm 1,\rm 2,*},
    JunGyu Lee\textsuperscript{\rm 3,*,$^{\ddagger}$},
    Gwanho Lee\textsuperscript{\rm 4},
    Jinwook Choi\textsuperscript{\rm 1,\rm 2},
    Young-Gon Kim\textsuperscript{\rm 2,\dag}
}
\affiliations{
    \textsuperscript{\rm 1}Seoul National University \quad
    \textsuperscript{\rm 2}Seoul National University Hospital \\
    \textsuperscript{\rm 3}Yonsei University \quad
    \textsuperscript{\rm 4}Korea University
}

\begin{document}

\renewcommand{\thefootnote}{\fnsymbol{footnote}}
\maketitle
\footnotetext[1]{These authors contributed equally.}
\footnotetext[2]{Corresponding author.}
\footnotetext[3]{Work done prior to joining Yonsei University.}

\begin{abstract} 

Accurate segmentation of thin curvilinear structures is vital for various real-world applications, from vessel analysis to road extraction. Yet their intricate geometry makes even minor pixel-wise errors enough to break the global topology, and this structural fragility turns severe domain shifts into catastrophic failures. The difficulty is most acute under cross-modality gaps, where the imaging process itself differs fundamentally between source and target. While test-time adaptation (TTA) offers a practical source-free remedy, existing methods adapt feature statistics and confidence, neither of which constrains connectivity, and thus degrade under such extreme gaps. To address this, we propose Skeleton-Guided Progressive Test-Time Adaptation (SGP-TTA). Progressive Batch Normalization (ProgBN) shifts normalization from frozen source statistics toward current target estimates under a sample-count schedule, so that the source–target balance follows the stage of adaptation rather than a fixed coefficient. Consensus Skeleton Recall (CSR) then derives a structural target from geometrically aligned multi-view predictions and updates only the BN affine parameters to preserve connected structures. Extensive experiments show that SGP-TTA consistently outperforms existing TTA methods in topological connectivity, with the largest margins under cross-modality shift. The project page is available at \url{https://boa-jang.github.io/SGP-TTA}.

\end{abstract}

\section{Introduction}

\begin{figure}[t]
    \centering
    \includegraphics[width=\columnwidth]{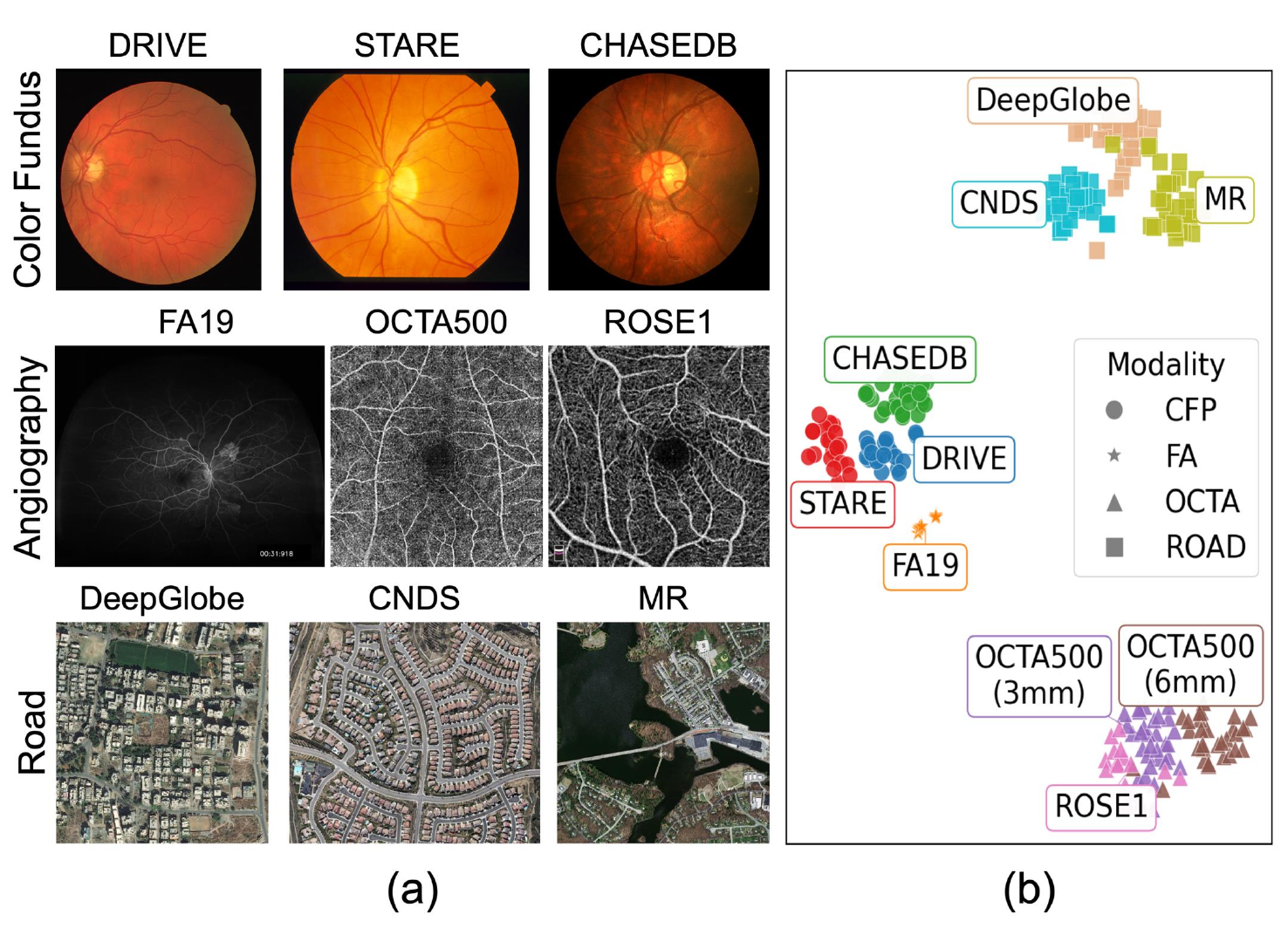}
    \vspace{-6mm}
    \caption{\textbf{Representative curvilinear structure datasets and their domain characteristics.} (a) Images from color fundus photography (CFP), OCT angiography (OCTA), fluorescein angiography (FA), and road datasets. (b) t-SNE of batch-normalization statistics; colors denote datasets, marker shapes denote imaging domains.}
    \label{fig1}
    \vspace{-2mm}
\end{figure}

Segmentation of thin curvilinear structures is a critical challenge in computer vision, arising in domains like satellite road extraction and vessel segmentation in medical imaging~\cite{bibiloni2016survey,cheng2021joint,goharbavang2025segmentation}. Because these narrow and intricately branched structures are highly susceptible to local prediction errors, minor misclassifications easily break structural connectivity. Such topological failures critically undermine downstream applications including autonomous navigation~\cite{bastani2018roadtracer} and biomarker quantification~\cite{giesser2024evaluating,montalt2022automatic}. Consequently, recent methods focus on preserving structural fidelity through topological architectures and connectivity losses~\cite{hu2019topology,shit2021cldice,kirchhoff2024skeleton}. Yet these methods rely heavily on supervision from a single domain, failing to generalize under inevitable distributional shifts. The limitation is most formidable across modalities. Retinal vasculature imaged by different systems and road networks captured by different sensors share the same underlying structure yet differ sharply in appearance, contrast, and background characteristics (Figure~\ref{fig1}).

To address severe domain shifts, unsupervised domain adaptation (UDA) typically exploits a pre-collected target dataset for offline alignment or retraining~\cite{hu2024adaptdiff,guo2026cross}. While effective, this requirement limits its applicability in deployment scenarios where target data are available only at inference time. Consequently, test-time adaptation (TTA) has emerged as a practical source-free alternative, adapting models on-the-fly during inference~\cite{wang2020tent,zhou2025topotta}. However, existing TTA frameworks are poorly matched to thin curvilinear structures for two reasons. First, normalization-based methods must determine how strongly to rely on target statistics estimated from a single test image. Immediately replacing the source statistics can destabilize early predictions because the batch normalization (BN) affine parameters remain tuned to source-normalized features. Conversely, a source-heavy mixture that stabilizes this early stage becomes increasingly conservative after the affine parameters have adapted. The appropriate source--target balance is therefore stage-dependent rather than constant. Second, confidence-based objectives such as entropy minimization are reduced as effectively by suppressing a weak branch as by confirming it, and under the extreme foreground–background imbalance of thin structures suppression is the cheaper direction; connectivity enters the objective nowhere and is therefore free to break.

To address these two limitations, we propose Skeleton-Guided Progressive Test-Time Adaptation (SGP-TTA), a unified source-free framework that separately adapts the normalization statistics and affine parameters of BN layers. This separation reflects their different roles during test-time adaptation. Normalization statistics can be recalibrated directly from the current target input, whereas affine parameters are updated through unsupervised optimization. SGP-TTA therefore combines Progressive Batch Normalization (ProgBN) for statistical recalibration with Consensus Skeleton Recall (CSR) for topology-aware parameter adaptation. 

ProgBN addresses the stage-dependent source--target trade-off by monotonically increasing the target-statistics weight as more target samples are processed. We use the number of observed samples as a simple, label-free surrogate for adaptation progress, allowing the model to remain anchored to source statistics early in the stream and to rely more strongly on current target estimates later. The resulting schedule is deterministic and requires neither offline coefficient learning nor stored target statistics. 

CSR addresses the lack of structural supervision in confidence-based adaptation. Predictions from geometrically transformed views are aligned and aggregated into a multi-view consensus map, from which a dilated skeleton is extracted as a self-supervised structural target. CSR encourages each view prediction to recover the consensus-supported centerline and, together with an entropy term, updates only the BN affine parameters. ProgBN and CSR thus play complementary roles, with ProgBN reducing statistical mismatch and CSR preserving the connectivity of thin foreground structures. Our contributions are threefold:

\begin{itemize}
    \item We identify a stage-dependent limitation in BN-statistics adaptation. A source--target mixture that stabilizes early predictions becomes overly conservative once the affine parameters have adapted. We address this limitation with ProgBN, a deterministic sample-count schedule that requires neither target labels nor offline coefficient learning.
    
    \item We introduce CSR, a self-supervised structural objective that aggregates geometrically aligned multi-view predictions into a consensus map, extracts a dilated skeleton from it, and encourages each view prediction to recover the consensus-supported centerline, supplying the connectivity constraint that confidence-based objectives lack.
    
    \item We combine the two into SGP-TTA, a source-free framework that adapts the normalization statistics and the affine parameters of BN through separate mechanisms.
\end{itemize}

\section{Related Work} 
\subsection{Curvilinear Structure Segmentation} 

Curvilinear structures are characterized by thin, elongated, and densely branched geometry, demanding both fine pixel-level accuracy and the preservation of global connectivity~\cite{bibiloni2016survey,cheng2021joint}. One line of work injects geometric and connectivity priors into the network architecture, where deformable and orientation-aware convolutions model local curvature and long-range dependencies~\cite{mou2021cs2,qi2023dynamic}, while connectivity- and point-set representations explicitly encode structural relations and irregular topologies~\cite{qin2019airwaynet,wang2022pointscatter}. A complementary line designs topology-aware objectives that penalize disconnected components and broken centerline overlooked by region-based losses~\cite{hu2019topology,clough2020topological}. In particular, clDice~\cite{shit2021cldice} and skeleton recall loss~\cite{kirchhoff2024skeleton} promote connectivity by aligning predictions with soft or ground-truth skeletons. More recently, adapter-based approaches built on segmentation foundation models have advanced domain-general curvilinear segmentation across vessels, roads, and other thin structures~\cite{zhu2026dual}. However, these methods require reference masks or structural annotations during training, limiting their applicability to adaptation on unlabeled target images at test time. 

\begin{figure*}[t]
    \centering
    \includegraphics[width=\textwidth]{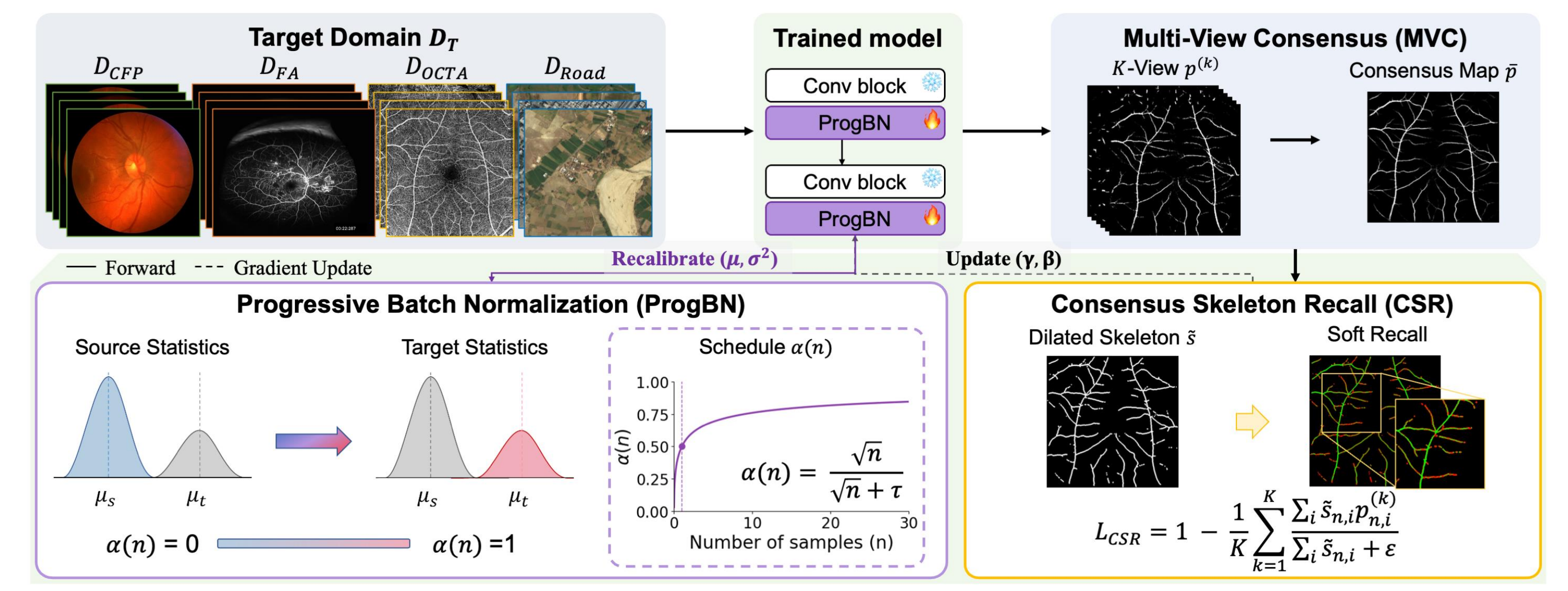}
    \vspace{-4mm}
    \caption{\textbf{Overview of SGP-TTA.} A source-trained network with frozen convolutions adapts only at its BN layers, through two complementary mechanisms. ProgBN blends frozen source and current target statistics under a sample-count schedule, without gradient updates, while CSR recalls the skeleton of the multi-view consensus map and updates only the BN affine parameters.} 
    \label{fig2}
\end{figure*}

\subsection{Cross-Modality Domain Adaptation} 

To bridge distribution shifts, UDA aligns source and target domains through adversarial feature alignment~\cite{ganin2016domain,tsai2018learning}, statistic matching~\cite{sun2016deep}, entropy-based adaptation~\cite{vu2019advent}, or image-to-image translation~\cite{hoffman2018cycada}. In the source-free setting, where source data is inaccessible, adaptation instead relies on pseudo-labeling and the intrinsic feature structure~\cite{chidlovskii2016domain,liang2020we}. For retinal vessels specifically, cross-modality methods synergistically align image and feature spaces~\cite{chen2019synergistic,chen2020unsupervised}, disentangle style from structure~\cite{peng2022unsupervised}, or synthesize target-style images via diffusion to fine-tune the segmenter~\cite{hu2024adaptdiff,zhang2025diffuseg}. Closer to our setting, topology-aware road adaptation predicts road skeletons as a domain-invariant auxiliary task and refines pseudo-labels by connectivity~\cite{iqbal2023leveraging}, yet still depends on source data and offline adversarial alignment. In general, these approaches require pre-collected target images and computationally heavy offline retraining, which is often impractical in unconstrained deployment.

\subsection{Test-Time Adaptation} 

TTA adapts a source-trained model to unlabeled targets on the fly, updating parameters with unsupervised surrogates such as entropy minimization~\cite{wang2020tent,niu2022efficient,niu2023towards}, continual self-training~\cite{wang2022continual,dobler2023robust}, or auxiliary self-supervision~\cite{sun2020test,liu2021ttt++}. Such objectives are fragile for thin structures, where the extreme foreground--background imbalance lets entropy minimization and pseudo-labeling reinforce errors around weak branches, and aggregating augmented views~\cite{khurana2021sita,zhang2022memo} reduces variability without preserving connectivity. Medical and curvilinear variants add visual prompts~\cite{chen2024each}, gradient alignment~\cite{chen2025gradient}, or topology-aware adaptation~\cite{zhou2025topotta}; the last supervises predictions with pseudo-labels placed in synthesized pseudo-breaks, so its structural signal is only as reliable as the output it perturbs, and it retains standard normalization. CSR instead reinforces the structure supported by the multi-view consensus rather than the structure the model is already confident about, and requires no ground-truth skeleton. 

Normalization-based TTA recalibrates BN statistics directly, replacing source statistics with target estimates~\cite{li2016revisiting,nado2020evaluating} or interpolating the two with a fixed coefficient~\cite{schneider2020improving,you2021test}; later works estimate the interpolation online, momentum-update toward the target, or robustify it against outliers and small batches~\cite{hu2021mixnorm,lim2023ttn,mirza2022norm,park2024medbn,kang2024membn}. Yet small-batch statistics remain unreliable under severe covariate shift~\cite{su2024unraveling}, suppressing the delicate foreground of thin structures. In contrast, ProgBN blends online target estimates with the frozen source statistics via a closed-form, monotone sample-count schedule, with no accumulated statistics or offline coefficient learning.

\section{Method}
\subsection{Preliminaries}
\subsubsection{Problem Setup.} 
We adapt a binary curvilinear-structure segmentation model $f_\theta$, trained on a labeled source domain $\mathcal{D}_S$, to an unlabeled target domain $\mathcal{D}_T$. The target is presented as a stream $\mathcal{D}_T=\{x_n\}_{n=1}^{N}$, where $n$ indexes the current target sample. The source and target domains share the same segmentation task but may differ substantially in image appearance, contrast, background characteristics, and acquisition conditions. Adaptation is source-free and label-free. Only the source-trained model is available, while the source data and target annotations are inaccessible. For a target image $x$, the model outputs a pixel-wise foreground probability map $p = \operatorname{sigmoid}(f_{\theta}(x))$ and a binary prediction $\hat{y}_i = \mathbf{1}\left[p_i \geq \delta\right]$, where $i$ indexes the spatial locations and $\delta$ is the prediction threshold. 

\subsubsection{Batch Normalization.} 
Every BN layer applies the same operation, so we describe a single layer and omit the layer index. Given an input feature map $h$, the layer normalizes it using the channel-wise statistics $(\mu,\sigma^2)$ and applies learnable affine parameters $(\gamma,\beta)$:
\begin{equation}
    \hat{h} = \frac{h-\mu}{\sqrt{\sigma^2+\epsilon}},
    \qquad y = \gamma\hat{h}+\beta,
\end{equation}
where $\epsilon$ is a small constant for numerical stability. At inference, BN normalizes the input using the running source statistics accumulated during training rather than statistics estimated from the test input. Under domain shift both components may be miscalibrated for the target, yet only the statistics are directly re-estimable from unlabeled target activations, whereas the affine parameters require a surrogate unsupervised objective.

\subsubsection{Overview.}

We propose Skeleton-Guided Progressive Test-Time Adaptation (SGP-TTA), which adapts the two BN components through separate mechanisms (Figure~\ref{fig2}). Progressive Batch Normalization (ProgBN) recalibrates the normalization statistics without gradient updates, while a single gradient step per image updates the affine parameters under a Consensus Skeleton Recall (CSR) loss and an entropy term. All non-BN weights and the source statistics remain frozen.

\subsection{Progressive Batch Normalization (ProgBN)}

Because the affine parameters remain calibrated to source-normalized features and are corrected only through successive gradient updates, replacing the source statistics immediately may destabilize early predictions, whereas a fixed source-heavy mixture becomes increasingly conservative once the affine parameters have adapted. ProgBN addresses this stage-dependent trade-off by progressively increasing its reliance on target statistics as the stream proceeds.

The same operation is applied to every BN layer, so we omit the layer index. Let $(\mu_s,\sigma_s^2)$ denote the frozen source running statistics, and let $n$ index the current target image, starting from $n=1$. For a target image $x_n$, let $(\mu_t(x_n),\sigma_t^2(x_n))$ denote the channel-wise statistics of its feature map $h_n$, computed over the spatial locations of the current image. These statistics carry no gradient and are discarded after processing the sample. ProgBN constructs the normalization statistics as
\begin{equation}
    \begin{aligned}
        \tilde{\mu}_n &= \alpha(n)\,\mu_t(x_n) + \bigl(1-\alpha(n)\bigr)\mu_s, \\
        \tilde{\sigma}_n^2 &= \alpha(n)\,\sigma_t^2(x_n) + \bigl(1-\alpha(n)\bigr)\sigma_s^2,
    \end{aligned}
\end{equation}
where $\alpha(n)\in(0,1)$ controls the reliance on the current target statistics. Following prior BN-statistics interpolation methods~\cite{schneider2020improving, you2021test}, the source and target means and variances are interpolated component-wise.

We use the number of processed target samples as a simple, label-free proxy for the stage of affine adaptation. To operationalize the transition from source-anchored to target-oriented normalization, we adopt a bounded square-root schedule that increases rapidly early in the stream while changing more gradually as adaptation proceeds:
\begin{equation}
    \alpha(n) = \frac{\sqrt{n}}{\sqrt{n}+\tau}, 
    \qquad 
    1-\alpha(n) = \frac{\tau}{\sqrt{n}+\tau},
\end{equation}
where $\tau>0$ controls the transition horizon. The schedule is monotonically increasing, the source contribution remains positive for every finite $n$, and no target labels are required. The source weight admits a pseudo-count reading: the frozen source statistics act as a prior worth $\tau$ units of adaptation progress, while $\sqrt{n}$ measures the progress made after $n$ target images. The count is not a measure of statistical precision: ProgBN never pools statistics across images, since each estimate is computed from the current image and discarded, so additional images sharpen the stage of affine adaptation rather than the statistic itself. That stage advances through one noisy gradient step per image; across $n$ steps the systematic component accumulates in proportion to $n$ while the fluctuations accumulate in proportion to $\sqrt{n}$, so its evidence grows as their ratio, $\sqrt{n}$. Thus $\tau$ carries a concrete meaning rather than serving as an opaque tuning constant, and the half-transition point $n = \tau^{2}$ follows immediately.

For an individual BN layer, ProgBN normalizes the current feature map using the progressively mixed statistics:
\begin{equation}
    \operatorname{ProgBN}_n(h_n) = \gamma_n \frac{h_n-\tilde{\mu}_n}{\sqrt{\tilde{\sigma}_n^2+\epsilon}} + \beta_n,
\end{equation}
where $\epsilon$ is a small constant. Among the model parameters, only the BN affine parameters $(\gamma_n, \beta_n)$ are updated across target samples. After the update on $x_n$, the complete set advances from $\Phi_n$ to $\Phi_{n+1}$, and we denote the model before this update as $f_n = f(\cdot\,;\Phi_n)$.

\begin{table*}[!t]
    \centering
    \caption{\textbf{Comparison with TTA methods for curvilinear structure segmentation under within-modality shifts across retinal imaging datasets.} Each column denotes a transfer setting. Source denotes the source model tested on the target domain without adaptation. The best and second-best results in each column are highlighted in \textbf{bold} and \underline{underline}, respectively.}
    \vspace{-2mm}
    \label{tab1}
    {\fontsize{8}{10}\selectfont
    \setlength{\tabcolsep}{4pt} 
    \renewcommand{\arraystretch}{1.15}
    \resizebox{\textwidth}{!}{
    \begin{tabular}{lcccccccccccc}
    \toprule
    \multicolumn{1}{c}{\textbf{Method}} 
    & \multicolumn{2}{c}{\textbf{DRIVE $\rightarrow$ STARE}}
    & \multicolumn{2}{c}{\textbf{DRIVE $\rightarrow$ CHASEDB}}
    & \multicolumn{2}{c}{\textbf{OCTA3mm $\rightarrow$ ROSE1}}
    & \multicolumn{2}{c}{\textbf{OCTA6mm $\rightarrow$ ROSE1}}
    & \multicolumn{2}{c}{\textbf{ROSE1 $\rightarrow$ OCTA3mm}} 
    & \multicolumn{2}{c}{\textbf{ROSE1 $\rightarrow$ OCTA6mm}} \\
    \cmidrule(lr){2-3}\cmidrule(lr){4-5}\cmidrule(lr){6-7}\cmidrule(lr){8-9}\cmidrule(lr){10-11}\cmidrule(lr){12-13}
    \multicolumn{1}{c}{}
    & Dice & clDice & Dice & clDice & Dice & clDice & Dice & clDice & Dice & clDice & Dice & clDice \\
    \midrule
    Source
    & 51.32 & 48.55 & 37.08 & 38.95 & 47.47 & 41.75 & 56.49 & 49.72 & 59.76 & 64.22 & 68.92 & 80.51 \\
    TENT~\cite{wang2020tent}
    & 70.96 & 68.66 & 71.60 & 73.51 & 48.47 & 41.65 & 53.77 & 48.79 & \underline{63.17} & \underline{66.07} & \underline{71.71} & \underline{80.89} \\
    CoTTA~\cite{wang2022continual}
    & 71.67 & 69.66 & 71.27 & 73.43 & 47.54 & 42.75 & 53.94 & 48.95 & 62.26 & 64.52 & 71.41 & 80.30 \\
    SAR~\cite{niu2023towards}
    & 70.96 & 68.69 & 71.37 & 73.22 & 48.55 & 41.73 & 53.84 & 48.87 & 62.17 & 64.44 & 71.36 & 80.26 \\
    EATA~\cite{niu2022efficient}
    & 71.06 & 68.78 & 71.61 & 73.56 & 48.57 & 43.73 & 53.87 & 49.79 & 62.44 & 65.05 & 71.42 & 80.60 \\
    MedBN~\cite{park2024medbn}
    & 70.26 & 68.15 & 70.13 & 71.33 & \underline{50.19} & \underline{43.89} & \underline{56.54} & \underline{51.81} & 49.86 & 50.83 & 61.76 & 71.54 \\
    VPTTA~\cite{chen2024each}
    & 70.92 & 68.03 & 71.33 & 73.47 & 49.23 & 42.75 & 56.05 & 51.06 & 54.43 & 55.75 & 67.97 & 78.44 \\
    GraTa~\cite{chen2025gradient}
    & 70.96 & 68.69 & 71.41 & 73.26 & 48.56 & 41.74 & 53.85 & 48.89 & 62.37 & 64.66 & 71.44 & 80.33 \\
    TopoTTA~\cite{zhou2025topotta}
    & \underline{71.98} & \underline{70.66} & \underline{71.70} & \underline{76.12} & 49.44 & 43.35 & 55.15 & 48.57 & 62.18 & 65.55 & 65.76 & 79.23 \\
    \rowcolor{lightblue}
    \textbf{SGP-TTA (Ours)}
    & \textbf{74.17} & \textbf{71.36}
    & \textbf{73.79} & \textbf{78.30}
    & \textbf{58.72} & \textbf{52.02}
    & \textbf{62.71} & \textbf{58.41}
    & \textbf{63.55} & \textbf{66.30}
    & \textbf{72.56} & \textbf{81.95}
    \\
    \bottomrule
    \end{tabular}
    }}
\end{table*}

\begin{table*}[!t]
    \centering
    \begin{minipage}[t]{0.60\textwidth}
    \centering
    \caption{\textbf{Comparison with TTA methods under cross-modality shifts across retinal imaging datasets.}}
    \vspace{-2mm}
    \label{tab2}
    {\fontsize{7.5}{9}\selectfont
    \setlength{\tabcolsep}{3.2pt} 
    \renewcommand{\arraystretch}{1.12}
    \resizebox{\linewidth}{!}{
    \begin{tabular}{lcccccccc}
    \toprule
    \multicolumn{1}{c}{\textbf{Method}} 
    & \multicolumn{2}{c}{\textbf{DRIVE $\rightarrow$ FA19}}
    & \multicolumn{2}{c}{\textbf{DRIVE $\rightarrow$ OCTA3mm}}
    & \multicolumn{2}{c}{\textbf{DRIVE $\rightarrow$ OCTA6mm}}
    & \multicolumn{2}{c}{\textbf{DRIVE $\rightarrow$ ROSE1}} \\
    \cmidrule(lr){2-3}\cmidrule(lr){4-5}\cmidrule(lr){6-7}\cmidrule(lr){8-9}
    \multicolumn{1}{c}{}
    & Dice & clDice & Dice & clDice & Dice & clDice & Dice & clDice \\
    \midrule
    Source
    & 40.75 & 40.09 & 25.84 & 24.39 & 37.92 & 36.60 & 46.35 & 44.31 \\
    TENT~\cite{wang2020tent}
    & 41.58 & 41.84 & 30.14 & 33.90 & 54.67 & \underline{63.02} & 40.74 & 42.12 \\
    CoTTA~\cite{wang2022continual}
    & 41.42 & 41.92 & 29.48 & 33.07 & 52.64 & 60.59 & 40.62 & 42.03 \\
    SAR~\cite{niu2023towards}
    & 42.61 & 42.89 & 29.58 & 33.50 & 53.42 & 61.37 & 41.63 & 43.25 \\
    EATA~\cite{niu2022efficient}
    & 41.60 & 41.88 & 30.49 & \underline{34.01} & 54.91 & 62.98 & 40.80 & 42.18 \\
    MedBN~\cite{park2024medbn}
    & 39.97 & 41.18 & 27.08 & 29.60 & 48.62 & 54.90 & 41.86 & 43.00 \\
    VPTTA~\cite{chen2024each}
    & \underline{44.79} & \underline{45.08} & \underline{32.50} & 33.91 & \underline{55.13} & 59.38 & 44.30 & 44.90 \\
    GraTa~\cite{chen2025gradient}
    & 41.62 & 41.89 & 29.48 & 33.10 & 52.88 & 60.90 & 40.64 & 42.04 \\
    TopoTTA~\cite{zhou2025topotta}
    & 43.85 & 44.01 & 24.18 & 23.33 & 35.51 & 35.36 & \underline{46.68} & \underline{47.04} \\
    \rowcolor{lightblue}
    \textbf{SGP-TTA (Ours)}
    & \textbf{47.53} & \textbf{49.65}
    & \textbf{44.83} & \textbf{46.41}
    & \textbf{59.62} & \textbf{69.71}
    & \textbf{53.95} & \textbf{51.16} \\
    \bottomrule
    \end{tabular}
    }}
    
    \end{minipage}
    \hfill
    \begin{minipage}[t]{0.38\textwidth}
    \centering
    \caption{\textbf{Comparison with TTA methods under cross-dataset shifts across road datasets.}}
    \vspace{-2mm}
    \label{tab3}
    {\fontsize{7.5}{9}\selectfont
    \setlength{\tabcolsep}{3.2pt} 
    \renewcommand{\arraystretch}{1.12}
    \resizebox{0.98\linewidth}{!}{
    \begin{tabular}{lcccc}
    \toprule
    \multicolumn{1}{c}{\textbf{Method}} 
    & \multicolumn{2}{c}{\textbf{DeepGlobe $\rightarrow$ MR}}
    & \multicolumn{2}{c}{\textbf{DeepGlobe $\rightarrow$ CNDS}} \\
    \cmidrule(lr){2-3}\cmidrule(lr){4-5}
    \multicolumn{1}{c}{}
    & Dice & clDice & Dice & clDice \\
    \midrule
    Source
    & 37.23 & 44.94 & \textbf{85.56} & \underline{93.39} \\
    TENT~\cite{wang2020tent}
    & 37.18 & 46.05 & 62.67 & 77.06 \\
    CoTTA~\cite{wang2022continual}
    & 37.35 & 46.32 & 62.91 & 77.20 \\
    SAR~\cite{niu2023towards}
    & 37.35 & 46.30 & 62.94 & 77.22 \\
    EATA~\cite{niu2022efficient}
    & 37.26 & 46.28 & 63.11 & 77.41 \\
    MedBN~\cite{park2024medbn}
    & 35.27 & 42.64 & 67.26 & 82.41 \\
    VPTTA~\cite{chen2024each}
    & 37.67 & 46.85 & 71.52 & 84.63 \\
    GraTa~\cite{chen2025gradient}
    & 37.40 & 46.38 & 63.00 & 77.30 \\
    TopoTTA~\cite{zhou2025topotta}
    & \textbf{42.24} & \underline{53.51} & 83.03 & 91.28 \\
    \rowcolor{lightblue}
    \textbf{SGP-TTA (Ours)}
    & \underline{40.80} & \textbf{54.84}
    & \underline{85.27} & \textbf{94.30} \\
    \bottomrule
    \end{tabular}
    }}
    \end{minipage}
\end{table*}

\subsection{Consensus Skeleton Recall (CSR)}

ProgBN recalibrates feature statistics but does not explicitly constrain the spatial connectivity of thin structures. We therefore derive a self-supervised structural target from prediction agreement across geometric views. For $K$ invertible transformations $\{T_k\}_{k=1}^{K}$, the pre-update model $f_n$ is applied to each transformed input, and the predictions are mapped back to the original coordinate frame to form a multi-view consensus (MVC):

\begin{equation}
    \begin{aligned}
        p_n^{(k)} &= T_k^{-1}\!\left[ \operatorname{sigmoid}\left(f_n\!\left(T_k(x_n)\right)\right)\right],\\
        \bar{p}_n &= \frac{1}{K}\sum_{k=1}^{K}p_n^{(k)}.
    \end{aligned}
\end{equation}

All views share the same affine parameters $\Phi_n$ and stream coefficient $\alpha(n)$, so the stream index advances once per image, while the target BN statistics are estimated independently for each transformed input.

Since local centerline errors can disconnect entire branches, we convert the consensus into a tubed skeleton target following Skeleton Recall Loss~\cite{kirchhoff2024skeleton}:

\begin{equation}
    \tilde{s}_n = \operatorname{Dilate}\!\left( \operatorname{Skel}\!\left( \mathbf{1}\!\left[\bar{p}_n\geq\delta\right] \right),r\right),
\end{equation}
where $r$ is the dilation radius used to tolerate minor disagreement across views. Unlike the supervised formulation, the skeleton is derived from the model's own consensus rather than from ground truth. Since this construction is non-differentiable, $\tilde{s}_n$ is treated as a stop-gradient target.

CSR encourages each aligned view prediction to cover the consensus-supported skeleton:
\begin{equation}
    \mathcal{L}_{\mathrm{CSR}} = 1- \frac{1}{K} \sum_{k=1}^{K} \frac{\sum_i \tilde{s}_{n,i}p_{n,i}^{(k)}}{\sum_i\tilde{s}_{n,i}+\varepsilon},
\end{equation}
where $i$ indexes spatial locations. Within $\mathcal{L}_{\mathrm{CSR}}$, gradients flow only through $p_n^{(k)}$ to the BN affine parameters; when the consensus skeleton is empty, the term contributes no gradient. The objective is recall-only and does not by itself penalize over-segmentation, but the update is confined to the BN affine parameters of a frozen network and takes a single step per image, so predictions stay close to the source model and do not drift toward a trivial all-foreground solution.

\subsection{Overall Objective and Online Adaptation}
A pixel-wise entropy term $\mathcal{L}_{\mathrm{ent}}$ on $\bar{p}_n$ further encourages confident predictions, and the objective combines two terms:
\begin{equation}
    \mathcal{L}_{\mathrm{SGP\text{-}TTA}}
    = \lambda_{\mathrm{CSR}}\,\mathcal{L}_{\mathrm{CSR}}
    + \lambda_{\mathrm{ent}}\,\mathcal{L}_{\mathrm{ent}},
\end{equation}
where $\lambda_{\mathrm{CSR}}$ and $\lambda_{\mathrm{ent}}$ are weights. A single gradient step on $\mathcal{L}_{\mathrm{SGP\text{-}TTA}}$ updates the BN affine parameters, after which the consensus is recomputed to form the prediction.

\begin{figure*}[t]
    \centering
    \includegraphics[width=\textwidth]{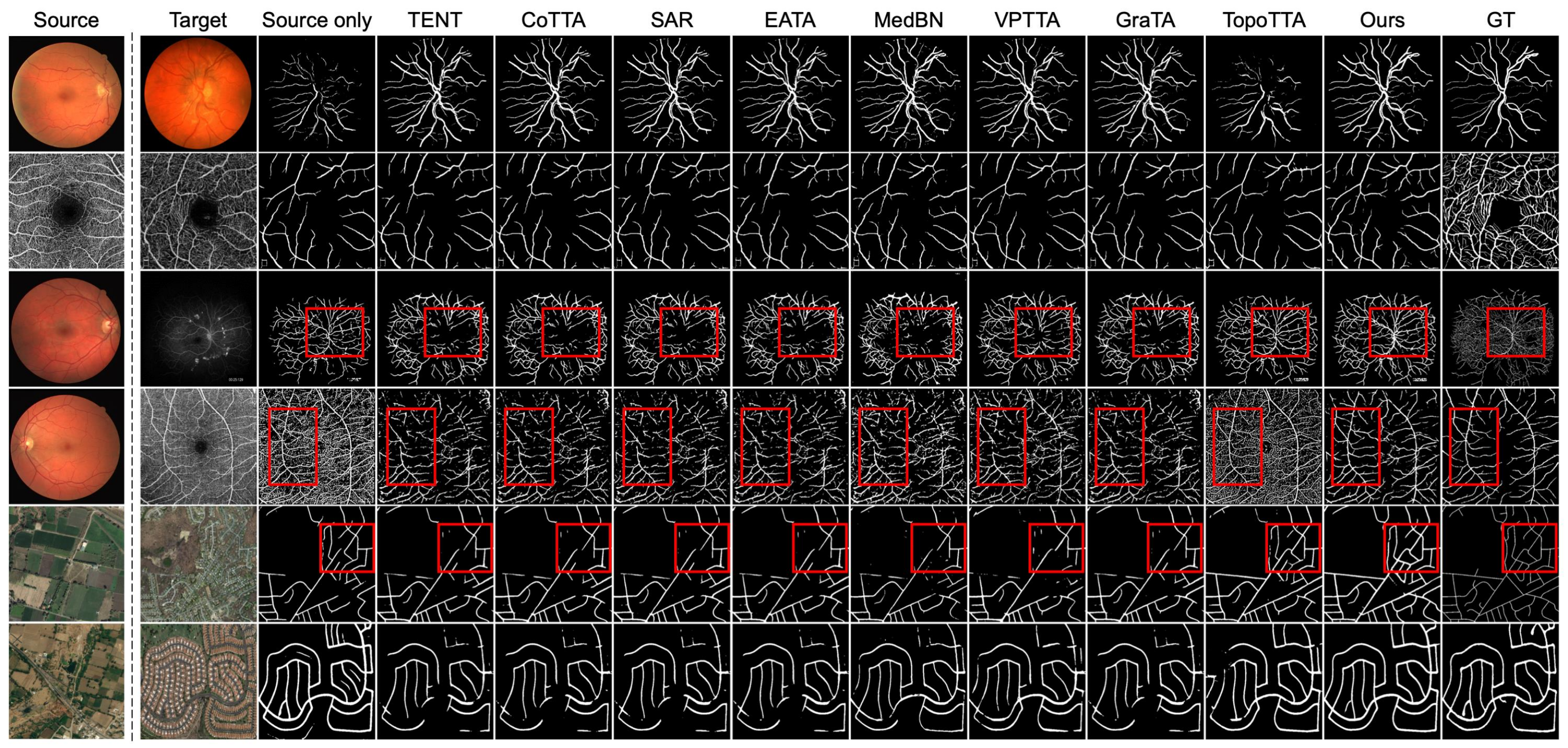}
    \vspace{-6mm}
    \caption{\textbf{Qualitative comparison across diverse target domains.} Red boxes indicate restored continuity of fine vessel structures and reduced false negatives in our SGP-TTA.}
    \label{fig3}
    \vspace{-2mm}
\end{figure*}

\begin{table*}[t]
    \centering
    \begin{minipage}[t]{0.55\textwidth}
    \centering
    \caption{\textbf{Comparison of BN adaptation strategies.} Each strategy replaces ProgBN in the source model, without MVC or CSR. Source and Target denote source-only and target-batch normalization.}
    \vspace{-2mm}
    \label{tab4}
    {\fontsize{8}{10}\selectfont
    \setlength{\tabcolsep}{4pt}
    \renewcommand{\arraystretch}{0.9}
    \resizebox{\linewidth}{!}{%
    \begin{tabular}{@{}l cc cc cc@{}}
    \toprule
    \multicolumn{1}{c}{\textbf{Method}} 
    & \multicolumn{2}{c}{\textbf{Retinal}}
    & \multicolumn{2}{c}{\textbf{Retinal (Cross)}}
    & \multicolumn{2}{c}{\textbf{Road}} \\
    \cmidrule(lr){2-3}\cmidrule(lr){4-5}\cmidrule(lr){6-7}
     & Dice & clDice & Dice & clDice & Dice & clDice \\
    \midrule
    Source
    & 53.51 & 53.95 & 37.72 & 36.35 & 61.40 & 69.17 \\
    Target
    & \underline{63.04} & \underline{62.87} & 41.01 & 44.32 & 50.15 & 61.76 \\
    \midrule
    AdaptiveBN~\cite{schneider2020improving}
    & 56.99 & 57.00 & 44.73 & 43.68 & \textbf{63.56} & 72.53 \\
    $\alpha$-BN~\cite{you2021test}
    & 61.40 & 60.95 & 43.19 & 45.93 & 61.32 & 71.98 \\
    MixNormBN~\cite{hu2021mixnorm}
    & 56.34 & 56.13 & 46.13 & 45.59 & \underline{63.48} & \underline{72.63} \\
    TTN~\cite{lim2023ttn}
    & 61.17 & 60.70 & \underline{46.62} & 45.45 & 61.29 & 71.97 \\
    MemBN~\cite{kang2024membn}
    & 60.38 & 59.95 & 46.21 & \underline{47.08} & 55.02 & 66.40 \\
    \rowcolor{lightblue}
    \textbf{ProgBN}
    & \textbf{64.12} & \textbf{63.76} & \textbf{47.07} & \textbf{48.13} & 62.80 & \textbf{72.75} \\
    \bottomrule
    \end{tabular}%
    }}
    \end{minipage}
    \vspace{-2mm}
    \hfill
    \begin{minipage}[t]{0.425\textwidth}
    \centering
    \caption{\textbf{Ablation of the proposed components.} MVC and ProgBN are first applied alone, then combined with each loss term.}
    \vspace{-2mm}
    \label{tab5}
    {\fontsize{8}{10}\selectfont
    \setlength{\tabcolsep}{4pt}
    \renewcommand{\arraystretch}{1.23}
    \resizebox{\linewidth}{!}{%
    \begin{tabular}{@{}l cc cc cc@{}}
    \toprule
    \multicolumn{1}{c}{\textbf{Configuration}} 
    & \multicolumn{2}{c}{\textbf{Retinal}}
    & \multicolumn{2}{c}{\textbf{Retinal (Cross)}}
    & \multicolumn{2}{c}{\textbf{Road}} \\
    \cmidrule(lr){2-3}\cmidrule(lr){4-5}\cmidrule(lr){6-7}
     & Dice & clDice & Dice & clDice & Dice & clDice \\
    \midrule
    Source
    & 53.51 & 53.95 & 37.72 & 36.35 & 61.40 & 69.17 \\
    MVC
    & 53.93 & 54.47 & 38.51 & 37.00 & 61.23 & 69.30 \\
    ProgBN
    & 64.12 & 63.76 & 47.07 & 48.13 & \underline{62.80} & 72.75 \\
    MVC + ProgBN
    & 65.22 & 65.10 & 47.84 & 51.42 & 59.15 & 70.62 \\
    MVC + ProgBN + $\mathcal{L}_{\mathrm{ent}}$
    & 62.36 & 62.02 & 35.24 & 38.71 & 44.98 & 53.08 \\
    MVC + ProgBN + $\mathcal{L}_{\mathrm{CSR}}$
    & \underline{67.53} & \underline{67.97} & \underline{49.70} & \underline{52.28} & 62.58 & \underline{73.75} \\
    \rowcolor{lightblue}
    \textbf{SGP-TTA}
    & \textbf{67.58} & \textbf{68.06} & \textbf{51.49} & \textbf{54.23} & \textbf{63.03} & \textbf{74.57} \\
    \bottomrule
    \end{tabular}%
    }}
    \end{minipage}
    \vspace{-2mm}
\end{table*}

\section{Experiments}
\subsection{Experiment Setup}

\subsubsection{Datasets and Evaluation Metrics.} 

We adopt 10 widely-used curvilinear segmentation datasets spanning four imaging domains. The retinal vessel datasets on color fundus photographs (CFP) are DRIVE~\cite{staal2004ridge}, STARE~\cite{hoover2000locating}, and CHASEDB~\cite{fraz2012ensemble}, containing 40, 20, and 28 images. For OCT angiography (OCTA), ROSE1~\cite{ma2020rose} and the two fields of view of OCTA500~\cite{li2024octa}, OCTA3mm and OCTA6mm, contain 117, 200, and 300 images. Fluorescein angiography (FA) is represented by RECOVERY-FA19~\cite{ding2020novel} (8 images). The road extraction datasets are DeepGlobe~\cite{demir2018deepglobe}, Massachusetts Roads (MR)~\cite{mnih2013machine}, and CNDS~\cite{cheng2017automatic}, with 8,570, 1,171, and 224 images. We follow the original splits and report Dice and clDice~\cite{shit2021cldice} for pixel-wise accuracy and topological continuity.

\subsubsection{Implementation Details.} 

For each transfer scenario, we use UNet~\cite{ronneberger2015u} as the primary backbone and resize all images to $512 \times 512$. The source model is trained on the labeled source domain with a combined Dice and cross-entropy loss, and every TTA method adapts from this same checkpoint. Because angiography renders vessels bright on a dark background while CFP renders them dark on a bright one, we invert the angiography images so that foreground polarity is consistent across domains; this uses only the known contrast polarity of each modality, not target annotations. The target domain is presented as a single-image stream. Images arrive one at a time, each is seen once, and predictions are binarized at $\delta=0.5$. 

For SGP-TTA, the MVC uses $K=6$ transformations, which include the identity, horizontal and vertical flips, and rotations of $90^\circ$, $180^\circ$, and $270^\circ$, which preserve the image grid. The consensus skeleton is dilated with a disk of radius $r=2$. We set $\lambda_{\mathrm{CSR}}=\lambda_{\mathrm{ent}}=1$ and take one Adam step per target image at learning rate $1\times10^{-3}$. The ProgBN parameter $\tau$ is $1$ for the retinal scenarios and $5$ for road extraction, whose target stream is more heterogeneous.

\subsubsection{Baselines.} 

SGP-TTA is evaluated at two levels. At the method level, it is compared against the source-only model (no adaptation) and eight test-time adaptation methods: TENT~\cite{wang2020tent}, CoTTA~\cite{wang2022continual}, SAR~\cite{niu2023towards}, EATA~\cite{niu2022efficient}, MedBN~\cite{park2024medbn}, VPTTA~\cite{chen2024each}, GraTa~\cite{chen2025gradient}, and TopoTTA~\cite{zhou2025topotta}. Since no official implementation of TopoTTA is available, we reproduce it following the descriptions in the original paper. At the normalization level, ProgBN is compared against five alternative BN strategies applied to the source model alone: AdaptiveBN~\cite{schneider2020improving}, $\alpha$-BN~\cite{you2021test}, MixNormBN~\cite{hu2021mixnorm}, TTN~\cite{lim2023ttn}, and MemBN~\cite{kang2024membn}. All baselines adapt under the same single-image stream with one gradient step per image; their hyperparameters follow the original papers.

\subsection{Comparison Results}

We evaluate SGP-TTA under three types of distribution shift: within-modality retinal transfer, cross-modality retinal transfer, and cross-dataset road transfer.

\subsubsection{Key Observation 1: Consistent performance across diverse shifts.}

SGP-TTA achieves the highest clDice in all 12 transfer settings and the highest Dice in 10 of 12. Under within-modality retinal shifts (Table~\ref{tab1}), the largest margin over existing baselines is on OCTA3mm $\rightarrow$ ROSE1, where SGP-TTA reaches 58.72 Dice, compared with 47.47 for the source model and 50.19 for the strongest baseline. The only settings in which SGP-TTA does not rank first in Dice are the two road transfers (Table~\ref{tab3}), where it nevertheless attains the highest clDice. On CNDS, the source model already achieves 85.56 Dice, leaving limited room for improvement. Whereas several TTA baselines reduce Dice to 62.67--71.52, SGP-TTA retains 85.27 Dice and raises clDice from 93.39 to 94.30. On MR, SGP-TTA obtains 40.80 Dice, slightly below the best of 42.24, but records the highest clDice of 54.84.

\subsubsection{Key Observation 2: Larger gains under cross-modality shifts.}

SGP-TTA shows its largest average gains in the cross-modality retinal setting (Table~\ref{tab2}). Several existing methods provide only limited improvements or degrade the source model in at least one transfer, whereas SGP-TTA improves both Dice and clDice over the source model across all four cross-modality settings. The largest margin is observed on DRIVE $\rightarrow$ OCTA3mm, where SGP-TTA achieves 44.83 Dice and 46.41 clDice, compared with the best baseline scores of 32.50 Dice and 34.01 clDice, respectively. Notably, TopoTTA, which also incorporates an explicit structural objective, performs below the source model on both OCTA500 transfers, whereas SGP-TTA consistently improves both metrics. This comparison indicates that reliable adaptation under severe modality shifts requires not only structural guidance but also stable feature-statistic recalibration. Figure~\ref{fig3} supports these quantitative trends. In the displayed examples, SGP-TTA recovers weak vessel branches and narrow road segments while leaving fewer visible discontinuities along the annotated structures.

\begin{figure}[t]
    \centering
    \includegraphics[width=1\columnwidth]{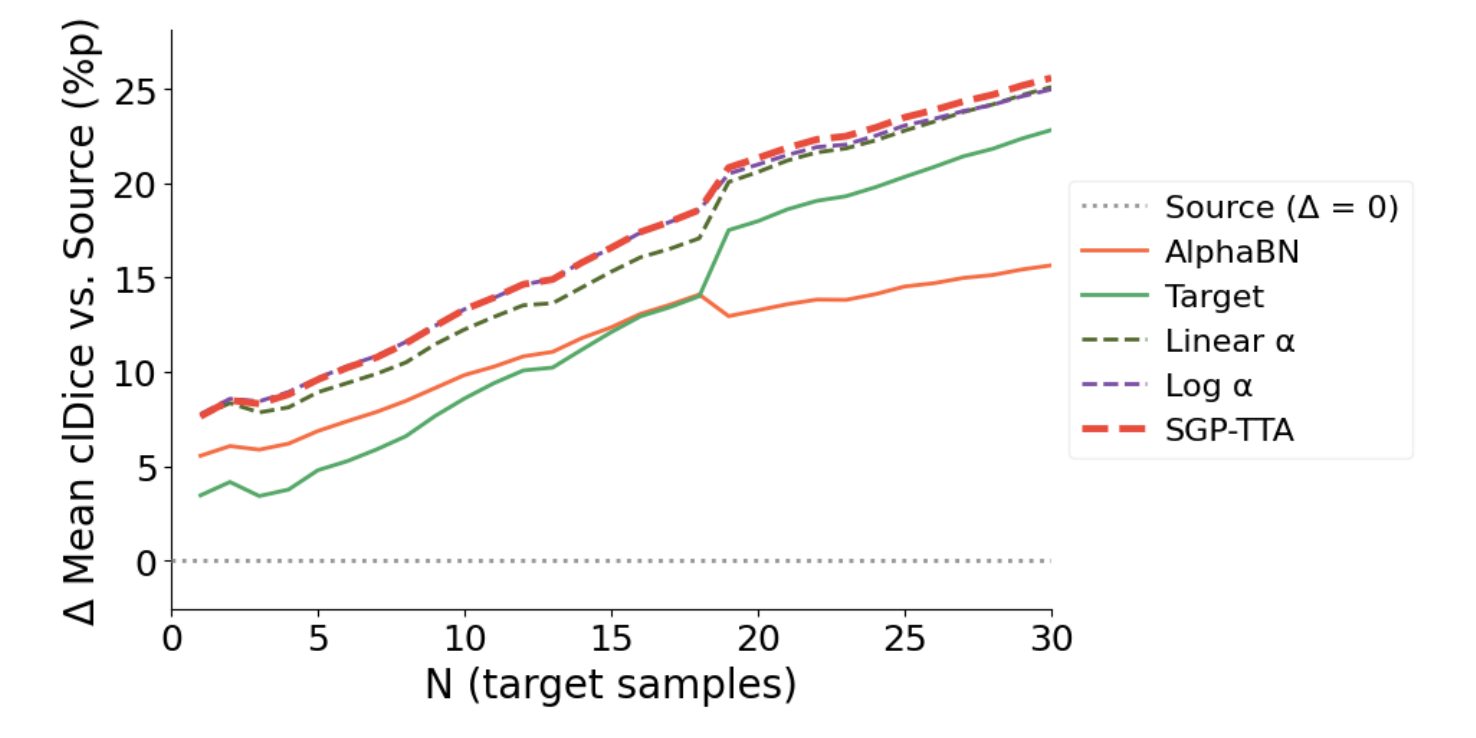}
    \caption{\textbf{Effect of the $\alpha$ schedule over the target stream.} Improvement over the source model against the number of observed target samples $N$, averaged over all settings.}
    \label{fig4}
    \vspace{-2mm}
\end{figure}

\subsection{Analysis and Ablation Studies}

\subsubsection{Analysis of BN Adaptation Strategies.} 

Table~\ref{tab4} isolates BN-statistics adaptation by removing MVC and gradient updates. No single source--target balance performs consistently. Target-only normalization is second only to ProgBN on within-modality retinal transfers, yet on road transfers it falls from 61.40 to 50.15 Dice against frozen source statistics, and neither endpoint is competitive under cross-modality shift. ProgBN instead adjusts the balance with the number of observed samples, achieving the highest clDice in all three groups and the highest Dice in both retinal groups.

\subsubsection{Choice of the $\alpha$ Schedule.} 

Figure~\ref{fig4} reports the improvement over the source model, removing the effect of image difficulty along the stream. No fixed coefficient remains effective throughout. Source-heavy normalization leads early, whereas target-only normalization starts lowest and overtakes it later in the stream. The linear schedule releases the source prior rapidly and the logarithmic one separates only late; the square-root schedule gives the largest improvement at every point. This reversal between the two fixed endpoints is direct evidence that the preferred balance shifts with the stage of affine adaptation, which no single coefficient can follow.

\subsubsection{Ablation of Adaptation Components.}

In Table~\ref{tab5}, MVC alone yields only marginal changes, whereas ProgBN provides the largest individual gain, indicating that statistical recalibration is the primary contributor. Adding MVC to ProgBN further improves both retinal groups, although its benefit does not transfer uniformly to the road setting. Entropy minimization without CSR degrades performance, reducing cross-modality Dice from $47.84$ to $35.24$, whereas CSR improves both metrics across all three groups.

\subsubsection{Robustness Across Segmentation Architectures.} 

\begin{figure}[t]
    \centering
    \includegraphics[width=1\columnwidth]{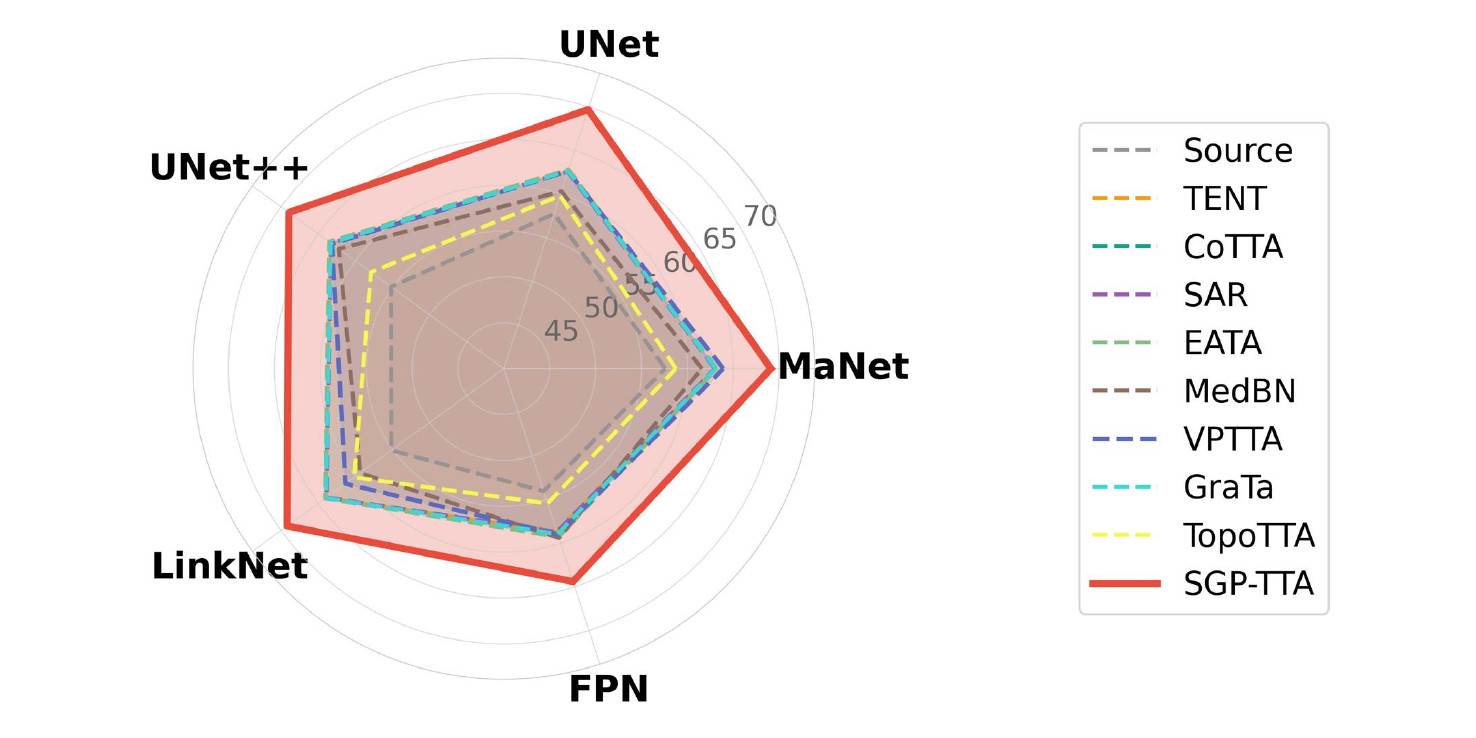}
    \caption{\textbf{Robustness across segmentation architectures.} Mean clDice on five different BN-based backbones, averaged over all settings.}
    \label{fig5}
    \vspace{-2mm}
\end{figure}

Figure~\ref{fig5} evaluates SGP-TTA across five BN-based backbones (UNet, UNet++, MaNet, LinkNet, and FPN), averaged over all transfer settings. SGP-TTA is the outermost method on every axis, so the gain does not rest on a particular decoder design and requires no architecture-specific modification. This is expected, since the method operates entirely on the BN layers these backbones share; by the same token, the claim extends to normalization-based networks and no further.

\subsubsection{Adaptation Cost.} 

\begin{figure}[t]
    \centering
    \includegraphics[width=1\columnwidth]{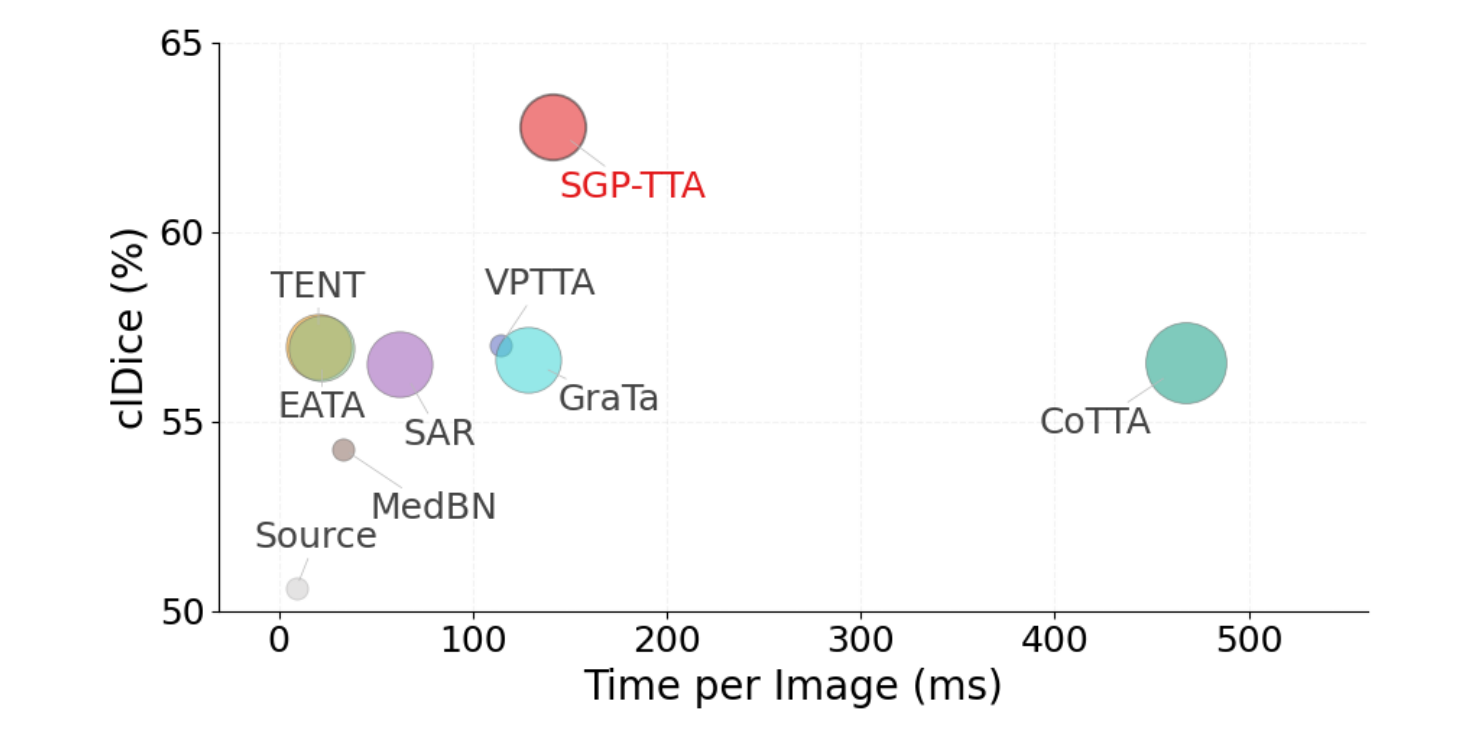}
    \caption{\textbf{clDice against adaptation cost.} Mean inference time per target image and clDice, averaged over all settings; marker area scales with the number of updated parameters.}
    \label{fig6}
    \vspace{-2mm}
\end{figure}

Figure~\ref{fig6} relates clDice to inference time and to the number of updated parameters. SGP-TTA is slower than the entropy-driven baselines, since it runs multiple views and a skeletonization step per image, but it remains well below the most expensive compared method while attaining the highest clDice. The entropy-driven methods cluster near the same clDice whether they update the BN affine parameters or the whole network, while SGP-TTA sits above them. The limiting factor therefore appears to be the adaptation objective rather than the compute spent on it.

\section{Conclusion} 
We presented SGP-TTA, a source-free test-time adaptation framework for thin curvilinear structures under single-image streams. Its premise is that the two domain-sensitive parts of a normalization layer fail differently and should be adapted differently. The statistics are recalibrated without gradients by a schedule over the number of images seen so far, while the affine parameters are trained against a skeleton consensus pooled over multiple views. The separation matters most under cross-modality transfer, where the target modality never appeared in training and existing methods degrade below the source model while ours improves both metrics. The useful mixture of source and target statistics moves as images accumulate, so no fixed proportion serves the whole stream; confidence is likewise a poor proxy for connectivity, so supervision must be anchored to the skeleton.

{\small
\bibliography{aaai2027}
}

\end{document}